\documentclass[conference]{IEEEtran}
\IEEEoverridecommandlockouts
\usepackage{cite}
\usepackage{amsmath,amssymb,amsfonts}
\usepackage{algorithmic}
\usepackage{graphicx}
\usepackage{textcomp}
\usepackage{xcolor}
\usepackage{multirow}
\usepackage[caption=false,font=footnotesize]{subfig}
\usepackage{pifont}
\def\BibTeX{{\rm B\kern-.05em{\sc i\kern-.025em b}\kern-.08em
    T\kern-.1667em\lower.7ex\hbox{E}\kern-.125emX}}

\begin{document}

\title{Assessing Covariate-Informed Grid Load Forecasting with a Time-Series Foundation Model}

\author{\IEEEauthorblockN{Varsha Pendyala,\IEEEmembership{~Member,~IEEE}, Yiwei Fu, Weizhong Yan,\IEEEmembership{~Senior Member,~IEEE}, and Nurali Virani}
\IEEEauthorblockA{\textit{Artificial Intelligence Team} \\
GE Vernova Advanced Research Center,
Niskayuna, NY, USA \\
\{varsha.pendyala, yiwei.fu, yan, nurali.virani\}@gevernova.com}
 }

\maketitle
\begin{abstract}
Modern power systems are growing increasingly complex as they integrate diverse generation sources to meet rising demand, making accurate load forecasting challenging. Recent advances in time-series foundation models (TSFMs) resulted in promising performance in zero-shot univariate load forecasting tasks. However, real-world load forecasting often involves multiple target variables and requires the integration of exogenous variables, raising important questions about the utility of TSFMs in realistic settings. In this study, we position Chronos-2, a recently developed model by Amazon, as a representative multi-channel TSFM that supports univariate, multivariate, and covariate-informed forecasting, and conduct a systematic investigation of how such models can be used for real-world load forecasting. While prior work has evaluated Chronos-2 on a limited number of energy-related tasks in a zero-shot setting, its performance relative to established task-specific deep learning models and its behavior when adapted using task-specific historical data remains insufficiently understood. In this work, we evaluate Chronos-2 on two real-world utility datasets, ISO New England and ENTSO-E, and benchmark it against widely used task-specific deep learning models. Our results show that Chronos-2 benefits substantially from task-specific fine-tuning and achieves strong short-horizon forecasting performance, but its zero-shot accuracy lags behind task-specific models and its forecasting error grows more rapidly with increasing forecast steps. Overall, this study provides a detailed characterization of the strengths and limitations of TSFMs such as Chronos-2 in grid load forecasting and offers practical insights into how a pretrained TSFM can be effectively adapted for operational load forecasting applications.
\end{abstract}


\section{Introduction}
Power grid load forecasting plays an important role in ensuring the reliable and economical supply of electricity for households, industries, and critical infrastructure. Accurate short-term (hours to days) and mid-term (weeks to months) load forecasting enables utilities to plan for efficient and cost-effective generation, perform maintenance operations, and procure fuel through economically-viable means~\cite{hong2016probabilistic}. As a result, load forecasting has long been a core problem in power system operations and planning. Deep learning approaches have been extensively studied for load forecasting, with applications ranging from  residential consumption prediction~\cite{masood2022multi,yan2018multi} to system-level demand forecasting~\cite{deng2019multi,fu2023masked,meghaa2025deep}. In recent years, the success of large-scale foundation models in natural language processing, computer vision, and speech has inspired the development of time-series foundation models (TSFMs)~\cite{goswami2024moment,das2024decoder,ansari2024chronos,rasul2023lag}. These models are typically pretrained on large collections of heterogeneous time series from multiple domains, including electricity, finance, and climate. Several recent studies have shown TSFMs achieving competitive performance as task-specific deep learning models on univariate load forecasting in zero-shot settings~\cite{meyer2024benchmarking,saravanan2024analyzing}.

However, real-world load forecasting is rarely univariate or isolated. Practical utility applications are often multivariate - such as forecasting demand simultaneously for multiple geographical regions and covariate-informed (e.g., requiring incorporation of exogenous variables such as weather and calendar features). To address these requirements, several TSFMs capable of processing multi-channel time series have recently been proposed. Moirai-1~\cite{woo2024unified} and Toto~\cite{cohen2025time} outperformed task-specific deep learning models on multivariate forecasting benchmarks in energy domain (ETT~\cite{mishra2025merged}, Electricity~\cite{electricityloaddiagrams20112014_321}), when past-covariates were allowed to be incorporated. COSMIC~\cite{auer2025zero} further considers covariates with known future values. It is shown that when covariates with both past and future values are available, traditional deep learning approaches performed better than zero-shot COSMIC. Many of these TSFMs remain limited in certain aspects crucial to load forecasting: lacking support for future covariates (Toto), restricted to single-target tasks (COSMIC), and high computational costs (Moirai-1).

More recently, Chronos-2~\cite{ansari2025chronos} was developed as a general-purpose TSFM that supports univariate, multivariate, and covariate-informed forecasting. While Chronos-2's performance was evaluated on several public benchmark datasets in zero-shot settings, it remains unclear how well it performs in multivariate and covariate-informed grid load forecasting and how it compares with established task-specific deep learning models. In this work, we address this gap through an in-depth evaluation of Chronos-2 on two real-world utility datasets, ISO New England~\cite{iso_ne_load_demand} and ENTSO-E~\cite{opsd_time_seris}, which involve multiple target variables along with diverse weather and calendar covariates. We benchmark Chronos-2 against widely used deep learning forecasting models, including temporal convolutional networks (TCN), long short-term memory (LSTM) networks, and transformer-based architectures. All the baseline models are trained using the Masked Multi-step Multivariate Forecasting (MMMF) framework~\cite{fu2023masked}, which has demonstrated strong performance on similar utility datasets.

While foundation models emphasize zero-shot generalization, most real-world utility applications have access to task-specific historical data. It is therefore important to understand how these pretrained models perform when they leverage such data. We evaluate Chronos-2 in zero-shot and adapted regimes and examine the trade-offs between full fine-tuning and parameter-efficient adaptation via LoRA~\cite{hu2022lora}. Further, we focus on two aspects that are central to operational load forecasting yet underexplored in prior TSFM studies. First, we examine how the calendar information can be incorporated when using pre-trained TSFMs, since it is critical for capturing seasonal demand patterns. Second, we analyze horizon-dependent error behavior to assess how forecasting error evolves across short-term and mid-term horizons that directly correspond to distinct grid operational decisions. Together, these analyses move beyond aggregate accuracy comparisons and provide practical guidance on how to effectively leverage pre-trained TSFMs for load forecasting and clarify the conditions under which their advantages over task-specific deep learning models can be realized.

Our main findings are: (1)~zero-shot Chronos-2 models exhibit substantially higher forecasting errors than task-specific deep learning models; (2)~fine-tuned Chronos-2 achieves significantly better short-horizon performance than deep learning models, with full fine-tuning outperforming LoRA-based adaptation; (3)~incorporating calendar variables during fine-tuning yields measurable performance improvements; and (4)~Chronos-2’s forecasting error increases more rapidly with each forecast step, whereas deep learning models display more stable error trajectories.

The remainder of the paper is organized as follows. Section~\ref{sec:background} provides an overview of Chronos-2 and the MMMF framework. Section~\ref{sec:eval_setup} describes the evaluation setup, datasets, Chronos-2 fine-tuning and baseline models. Section~\ref{sec:results} presents the experimental results and finally, Section~\ref{sec:conclude} concludes the paper by summarizing key insights and outlining directions for future work.

\section{Background}\label{sec:background}
\subsection{The Chronos-2 model}\label{sec:chronos-2}
Chronos-2~\cite{ansari2025chronos} is a 120M-parameter encoder-only Transformer model trained on a variety of univariate, multivariate, and covariate-informed forecasting tasks. Each series fed into the model represents either a target variable or a covariate. For a task with $D$ targets, $M$ covariates, a history length of $T$, and a forecast horizon $H$, the model input has dimensions $(D+M) \times (T+H)$. Categorical features are converted into numerical form using either target encoding or ordinal encoding.

All the input series to the model are first standardized and transformed using an inverse hyperbolic sine function. Meta features, including time index and a mask indicating observed or missing values, are included at each time step. The mask feature marks missing values for future targets and unknown future covariates, while known future covariates are marked as observed. Each series is then divided into non-overlapping patches, with each patch mapped to a fixed embedding size by a tokenizer and subsequently processed by the encoder. 

The model employs alternating time-attention and group-attention layers. Time-attention captures temporal dependencies within each series, while group-attention enables information sharing across series within a defined group. In a typical covariate-informed multivariate forecasting task, all targets and available covariates are assigned to the same group to capture cross-series dependencies and covariate–target interactions. Group-attention allows in-context learning (ICL) by enabling the model to share information across multiple time series, including targets, past-only covariates, and future-known covariates, by assigning them to a single group. In~\cite{ansari2025chronos}, it is shown that this mechanism improves forecasting performance compared to purely univariate modeling, with the largest gains observed for covariate-informed tasks.

Chronos-2 is trained using a quantile regression objective, producing direct multi-step forecasts for multiple quantiles. Pretraining is performed in two stages: in the first stage, a short forecast horizon is used; in the second stage, the forecast horizon is increased to 1,024. The exact forecast horizon used in the first stage is not specified in~\cite{ansari2025chronos}.

\subsection{Masked Multi-step Multivariate Forecasting (MMMF)}\label{sec:mmmf}
We employ Masked Multi-step Multivariate Forecasting (MMMF) framework~\cite{fu2023masked} as our primary baseline because prior work has demonstrated its superior performance on real-world utility datasets compared to traditional sample-based regression, recursive single-step forecasting (RSF), and direct multi-step forecasting (DMF) approaches. MMMF is a self-supervised learning framework that can be applied to any neural network architecture capable of processing sequential inputs. Unlike recursive methods that generate multi-step forecasts by feeding predictions back into the model thereby accumulating errors over the horizon, MMMF formulates forecasting as a masked reconstruction task that incorporates known future information.

Formally, let $\mathbf{x}_t$ represent covariates and $\mathbf{y}_t$ represent target variables at time $t$. The goal is to forecast targets for a horizon $H$ given a history $T$ and known future covariates (e.g., weather forecasts, calendar features). The model processes a sequence of length $(T+H)$. During training, the model receives the full sequence of covariates $\mathbf{x}_{t-T}, \dots, \mathbf{x}_{t+H}$, the historical targets $\mathbf{y}_{t-T}, \dots, \mathbf{y}_{t}$, and the segment of future target values $\mathbf{y}_{t+1}, \dots, \mathbf{y}_{t+H}$ that are masked and replaced with random values. The model is trained to reconstruct these masked values by leveraging both the temporal dependencies with the history and the explicit signals provided by the future covariates.

The loss function is calculated solely on the masked segments of the target variables. During inference, the desired forecast horizon is masked, and the model directly generates multi-step predictions while avoiding the error propagation inherent in recursive strategies.

\section{Evaluation Setup}\label{sec:eval_setup}
Chronos-2 is evaluated in zero-shot as well as after task-specific fine-tuning. It is fine-tuned under multiple configurations: forecasting using only the history of target variables versus forecasting that also depends on covariates (past and future), full fine-tuning versus Low Rank Adaptation (LoRA), and fine-tuning for different values of forecast horizon $H$. All the Chronos-2 and MMMF models are trained on an NVIDIA H200 GPU. During inference, every model is provided with a fixed history of 30 days. For Chronos-2, we always report MAPE with respect to the model's median prediction.

\subsection{Datasets}
Table~\ref{tab:datasets} shows the characteristics of the two real-world utility datasets used in this study.

\textbf{ISO New England} dataset~\cite{iso_ne_load_demand} provides electricity demand and weather data for eight zones in the New England region: Connecticut (CT), Maine (ME), Northeast Massachusetts and Boston (NEMA), New Hampshire (NH), Rhode Island (RI), Southeast Massachusetts (SEMA), Vermont (VT), and West/Central Massachusetts (WCMA). Following the experimental setup in~\cite{fu2023masked}, daily data from 2011 to 2020 are used for training or fine-tuning all the models, while the data from 2021 is reserved for evaluation. In this forecasting task, the target variables are the electricity demand for each zone, and the covariates include calendar features (month, date, and day of week) and zone-wise weather variables: dew point temperature and dry bulb temperature.

\textbf{ENTSO-E} dataset is obtained from Entsoe-1H multivariate electricity load forecasting task of the fev-bench benchmark~\cite{shchur2025fev}. Entsoe-1H combines electricity load data originally published by the European Network of Transmission System Operators for Electricity (ENTSO-E), accessed through the \texttt{time\_series} package~\cite{opsd_time_seris}, with the weather covariates from \texttt{weather\_data} package~\cite{opsd_weather} for six European countries: Austria (AT), Belgium (BE), Germany (DE), Hungary (HU), Luxembourg (LU), and the Netherlands (NL). We downsampled the original hourly data to daily frequency by taking the maximum values for each day. Data from 2015 to 2018 is used for training, and data from 2019 for evaluation. The target variables in this task are the daily peak electricity demand for each country, and the covariates are the calendar features and the country-wise weather variables: temperature, direct horizontal radiation and diffuse horizontal radiation.

\begin{table}[tb]
\caption{Datasets used in this study.}
\label{tab:datasets}
\centering
\renewcommand{\arraystretch}{1.2}
\begin{tabular}{p{1.5cm} p{2.8cm} p{1.4cm} p{1.2cm}}
\hline
{\textbf{Dataset}} &
{\textbf{Weather covariates}} &
{\textbf{Train split}} &
{\textbf{Test split}} \\
\hline
\hline
ISO‑New England
\newline (8 zones) &
Dew point and dry bulb temperature &
2011--2020 
\newline (3{,}653 days) &
2021 
\newline (365 days) \\
\hline
ENTSO-E
\newline (6 countries) &
Temperature; direct and diffuse horizontal radiation &
2015--2018 
\newline (1{,}461 days) &
2019 
\newline (365 days) \\
\hline
\end{tabular}
\end{table} 

\subsection{Chronos-2 fine-tuning}
 In each configuration, we fine-tune Chronos-2 models for 20 epochs, and the checkpoint corresponding to the best validation performance, measured by mean absolute percentage error (MAPE), is selected for evaluation on the test set. 20\% of the training set in each dataset is used for validation. Each model fine-tuning is repeated using five different random seeds, and the mean and standard deviation of the results are reported. The training uses AdamW optimizer with an initial learning rate of $10^{-6}$ for full fine-tuning and $10^{-5}$ for LoRA. Because LoRA learns a substantially smaller number of parameters, it could accommodate a higher learning rate without causing optimization divergence. A learning rate scheduler was used to reduce the learning rate by a factor of 0.5 whenever the validation metric did not improve for three consecutive epochs.

\subsection{Baseline models}
Chronos-2 is compared against these deep learning models trained using the MMMF framework~\cite{fu2023masked}: Temporal Convolutional Network (TCN), Long Short-Term Memory (LSTM), Transformer. Table~\ref{tab:baseline_models} provides the key hyperparameters and the size of these models. All these models use an embedding layer with 5 dimensions to encode categorical variables. Similar to Chronos-2, these MMMF models processes the target and covariate series corresponding to $(T+H)$ time steps, with unknown values masked. However, unlike Chronos-2, the input series are not segmented into patches and the MMMF does not rely on a fixed forecast horizon during the training. Instead, forecast horizon is randomly sampled for each mini-batch. We train all these models for 1000 epochs using Adam optimizer with a learning rate of 0.001.
\begin{table}[t]
\centering
\caption{Architecture details of MMMF models.}
\label{tab:baseline_models}
\renewcommand{\arraystretch}{1.2}
\begin{tabular}{c p{4.3cm} c}
\hline
\textbf{Model} & \textbf{Hyperparams} & \textbf{\# Params} \\
\hline
\hline
TCN & 2 conv layers, 50 channels, kernel: 3, stride: 1, dilation: $2^i$ ($i^\text{th}$ layer), dropout: 0.2 &  32,813\\
\hline
LSTM & 2 layers, hidden dim: 50 & 40,463 \\
\hline
Transformer & 2 encoder layers, model dim: 64, FF dim: 256, heads: 8 & 103,751 \\
\hline
\end{tabular}
\end{table}

\section{Results and Discussion}\label{sec:results}
Tables~\ref{tab:avg_mape_isone} and~\ref{tab:avg_mape_entsoe} report the average MAPE of Chronos-2 and MMMF-based baseline models trained with three different forecast horizons $H$, both with and without covariates. For Chronos-2, $H$ denotes the exact future length included in each input sample, whereas for MMMF models it corresponds to the maximum length of future targets masked during training. For each dataset, MAPE is averaged across all regions and over all forecast steps from day 1 to day $H$.

\subsection{Zero-shot vs. Fine-tuning}
Across both datasets, MMMF models outperform pre-trained Chronos-2 in the zero-shot setting. Moreover, the performance gap between MMMF models and zero-shot Chronos-2 increases as the forecast horizon $H$ grows, highlighting the limited zero-shot generalization of Chronos-2 in these energy datasets. Task-specific fine-tuning of Chronos-2 substantially improves forecasting accuracy. In many configurations, fine-tuned Chronos-2 models either outperform the MMMF baselines or achieve comparable performance. Full fine-tuning consistently yields better performance than LoRA-based fine-tuning, and this is particularly pronounced in settings with covariates. These results suggest that when sufficient task-specific data and informative covariates are available, full fine-tuning enables Chronos-2 to better adapt its representations to the downstream task, resulting in lower MAPE.

\begin{table}[tb]
\caption{Average MAPE of models trained for different forecast horizons over ISO New England$^*$.}
\label{tab:avg_mape_isone}
\centering
\renewcommand{\arraystretch}{1.2}
\begin{tabular}{c c c c c c}
\hline
    \textbf{Method} & \textbf{Covariates} & \textbf{Model type} & \textbf{H=30} & \textbf{H=60} & \textbf{H=90} \\ \hline \hline
    \multirow{3}{*}{MMMF} & \multirow{3}{*}{\ding{51}} & TCN & 5.51 & 5.72 & 5.77 \\  
    & & LSTM & 5.19 & 6.15 & 6.61 \\
    & & Transformer & \underline{4.13} & \textbf{5.06} & \underline{5.5} \\
    \hline     
    \multirow{3}{*}{Chronos-2} & \multirow{3}{*}{\ding{51}} & Zero-shot & 6.71 & 10.61 & 15.78 \\    
    & & LoRA fine-tune & 4.33 & 5.51 & 5.81 \\
    & & Full fine-tune & \textbf{4.06} & \underline{5.19} & \textbf{5.36} \\
    \hline \hline
    \multirow{3}{*}{MMMF} & \multirow{3}{*}{\ding{55}} & TCN & 13.26 & 14.49 & 14.78 \\  
    & & LSTM & 11.3 & \underline{11.61} & \underline{11.16} \\
    & & Transformer & 10.64 & \textbf{10.61} & \textbf{11.14} \\
    \hline     
    \multirow{3}{*}{Chronos-2} & \multirow{3}{*}{\ding{55}} & Zero-shot & 11.41 & 14.29 & 16.56 \\    
    & & LoRA fine-tune & \textbf{10.02} & 12.12 & 12.95 \\
    & & Full fine-tune & \underline{10.07} & 12.1 & 12.87 \\
    \hline  
\multicolumn{6}{l}{$^{*}$Best (per $H$, with/without covariates) in bold; second-best underlined.}
\vspace{-3mm}
\end{tabular}
\end{table}

\begin{table}[tb]
\caption{Average MAPE of models trained for different forecast horizons over ENTSO-E$^*$.}
\label{tab:avg_mape_entsoe}
\centering
\renewcommand{\arraystretch}{1.2}
\begin{tabular}{c c c c c c}
\hline
    \textbf{Method} & \textbf{Covariates} & \textbf{Model type} & \textbf{H=30} & \textbf{H=60} & \textbf{H=90} \\ \hline \hline
    \multirow{3}{*}{MMMF} & \multirow{3}{*}{\ding{51}} & TCN & \underline{4.52} & \textbf{4.51} & \textbf{4.43} \\  
    & & LSTM & 5.21 & 5.22 & 5.39 \\
    & & Transformer & 4.65 & 4.7 & \underline{4.82} \\
    \hline     
    \multirow{3}{*}{Chronos-2} & \multirow{3}{*}{\ding{51}} & Zero-shot & 5.63 & 6.57 & 7.67 \\    
    & & LoRA fine-tune & 4.54 & 4.91 & 5.32 \\
    & & Full fine-tune & \textbf{4.31} & \underline{4.66} & 4.86 \\
    \hline \hline
    \multirow{3}{*}{MMMF} & \multirow{3}{*}{\ding{55}} & TCN & 8.31 & 8.83 & 9.09 \\  
    & & LSTM & 5.66 & 6.5 & \textbf{7.12} \\
    & & Transformer & 6.95 & 7.71 & 8.2 \\
    \hline     
    \multirow{3}{*}{Chronos-2} & \multirow{3}{*}{\ding{55}} & Zero-shot & 5.78 & 6.59 & 7.53 \\    
    & & LoRA fine-tune & \underline{5.26} & \underline{6.25} & \underline{7.31} \\
    & & Full fine-tune & \textbf{5.16} & \textbf{6.16} & 7.36 \\
    \hline
\multicolumn{6}{l}{$^{*}$Best (per $H$, with/without covariates) in bold; second-best underlined.}
\vspace{-5mm}
\end{tabular}
\end{table}
\subsection{Calendar variables}
Since the original Chronos-2 implementation does not natively support calendar features, we employ different strategies to incorporate those features during the model fine-tuning. We examine the impact of two types of feature encoding to transform the calendar variables to numerical form for feeding to the Chronos-2 model. Those results (for $H=60$) are summarized in Table~\ref{tab:date_effect_isone}. In the first approach, ordinal encoding is first applied to the calendar variables and then normalized using min-max scaling. In the second approach, cyclical encoding is employed, where each calendar variable is represented using sine and cosine transformations with periods corresponding to the underlying temporal cycles. Because Chronos-2 applies instance normalization independently to each input time series, we explicitly disable it for calendar feature series. This is necessary to preserve meaningful temporal structure. For example, instance-normalizing the month feature over a 30-day history would eliminate seasonal information. The results show that calendar features do not improve performance in the zero-shot setting. In contrast, fine-tuned models are able to leverage calendar information effectively, with cyclical encoding yielding the best performance. This may be because cyclical encoding preserves the proximity of adjacent time points in the encoded space, unlike ordinal encoding, which places the beginning and end of a cycle (e.g., December and January) far apart.
\begin{table}[tb]
\caption{Ablation study over ISO New England on the impact of calendar feature encoding$^*$.}
\label{tab:date_effect_isone}
\centering
\renewcommand{\arraystretch}{1.2}
\begin{tabular}{c c c }
    \hline
    \textbf{Encoding} & \textbf{Zero-shot} & \textbf{{Full fine-tune}}\\ \hline \hline
    No date features & {10.15} & 5.58\\
    Ordinal & 10.33 & 5.3\\
    Cyclical & 10.61 & 5.19\\
    \hline
\multicolumn{3}{l}{$^{*}$Table shows average MAPE across horizon $H=60$.}
\vspace{-3mm}
\end{tabular}
\end{table}

\subsection{Performance at various horizons}
\begin{figure*}[tb]
\centering
\setlength{\tabcolsep}{2pt}
\begin{tabular}{c c c c c}
& \textbf{Chronos-2} & \textbf{MMMF-TCN} & \textbf{MMMF-LSTM} & \textbf{MMMF-Transformer} \\

\raisebox{3mm}{\rotatebox{90}{\textbf{ISO New England}}} &
\includegraphics[width=0.22\textwidth]{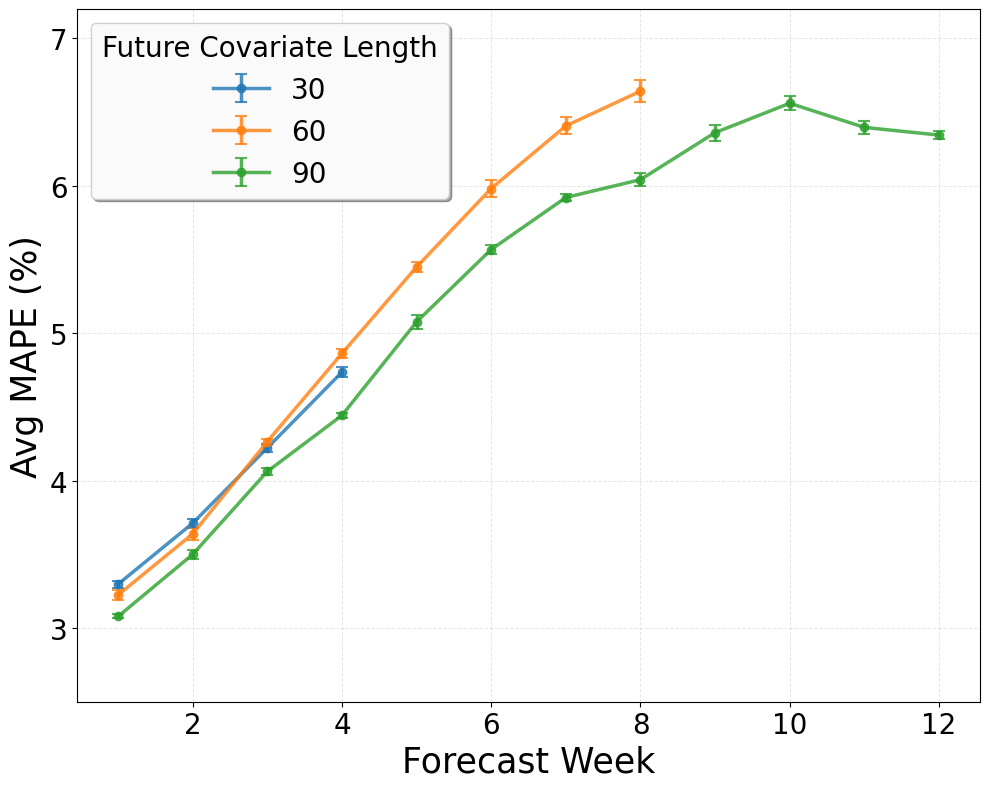} &
\includegraphics[width=0.22\textwidth]{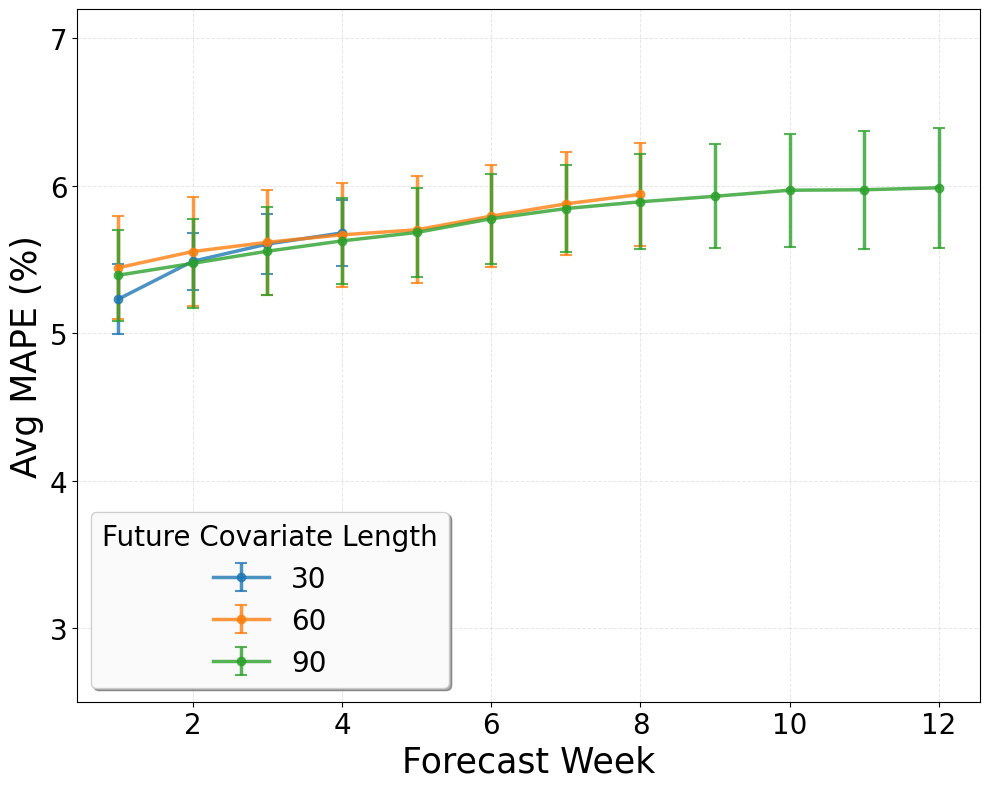} &
\includegraphics[width=0.22\textwidth]{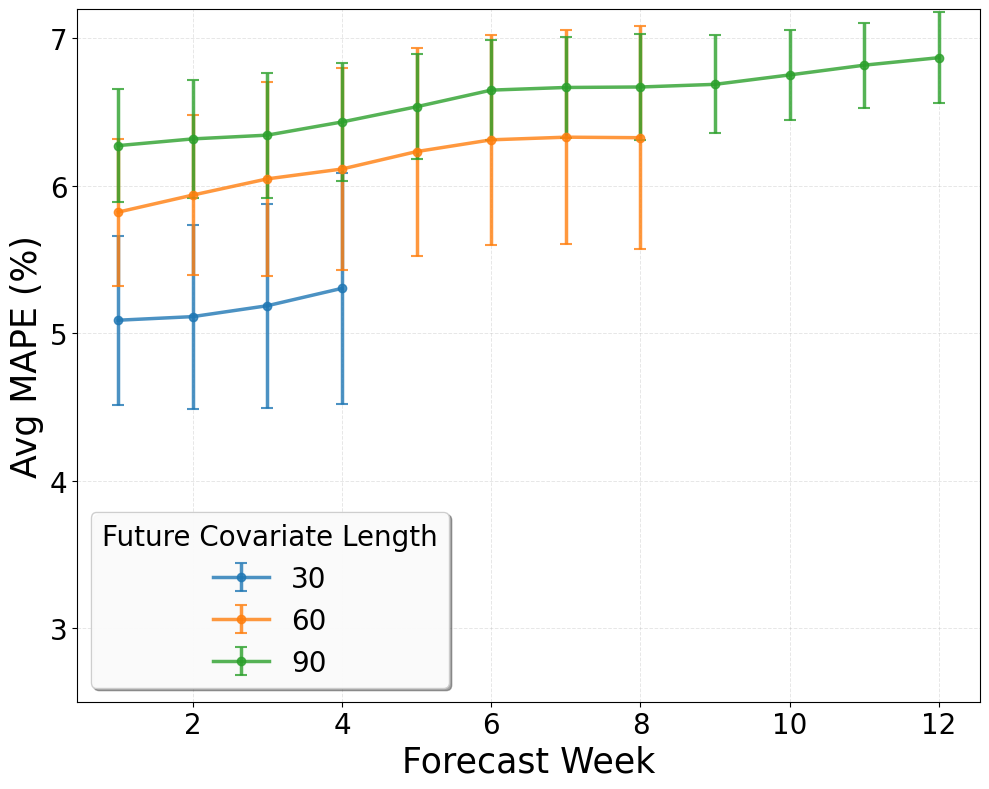} &
\includegraphics[width=0.22\textwidth]{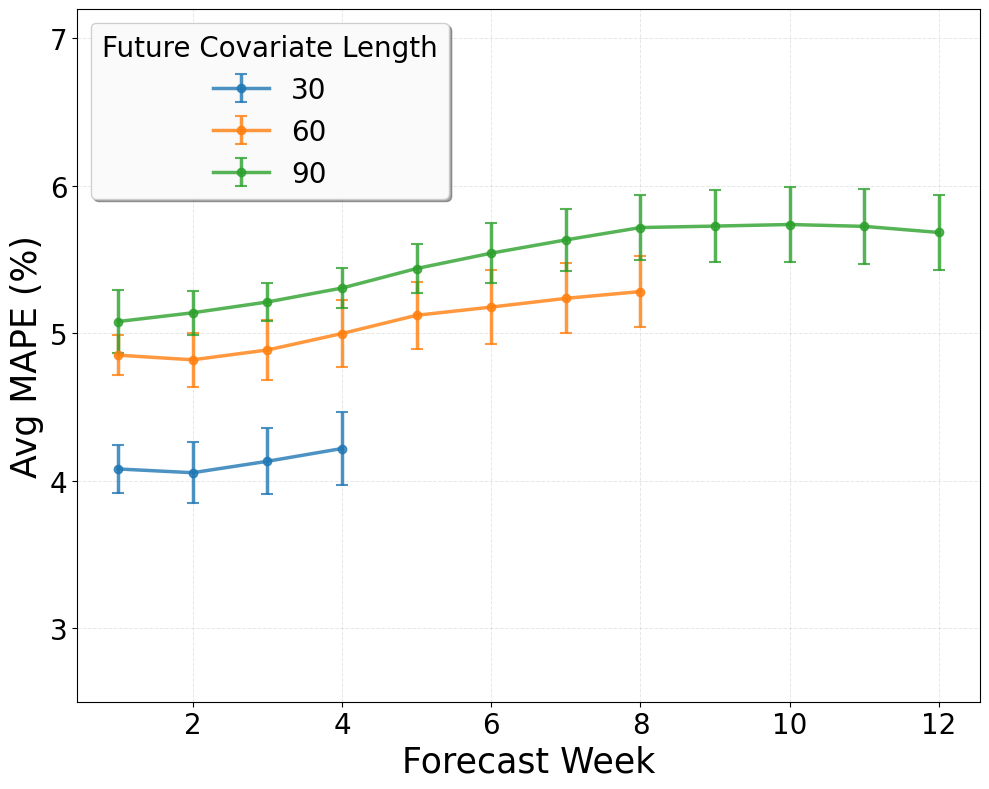} \\

\raisebox{8mm}{\rotatebox{90}{\textbf{ENTSO-E}}} &
\includegraphics[width=0.22\textwidth]{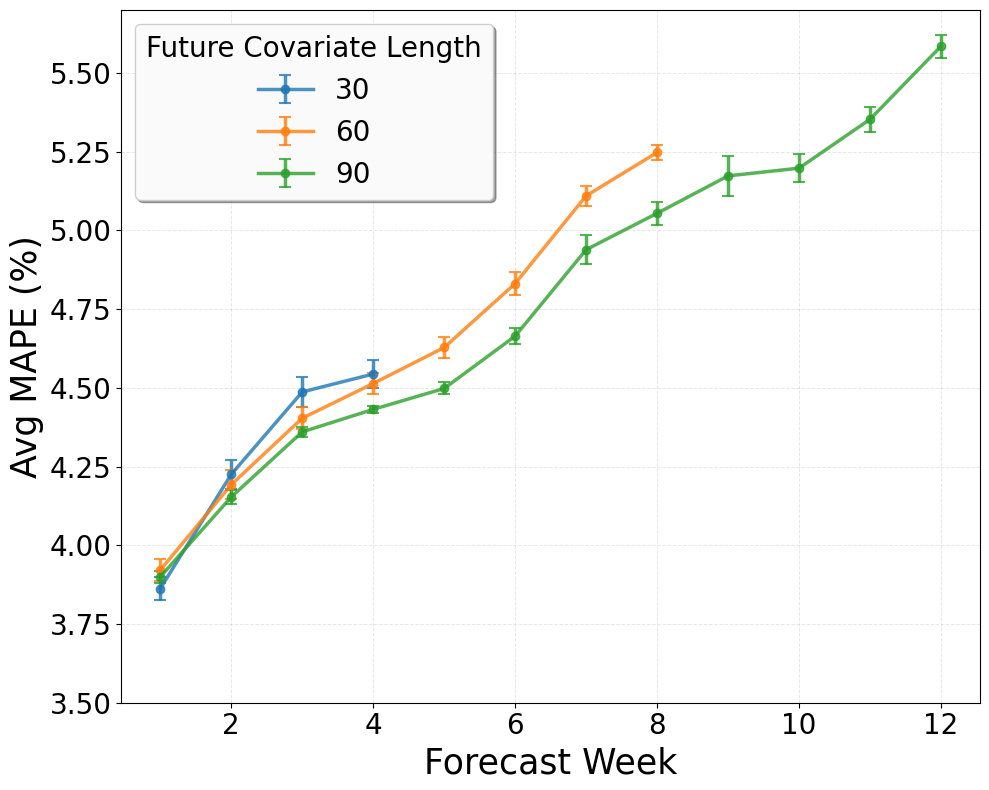} &
\includegraphics[width=0.22\textwidth]{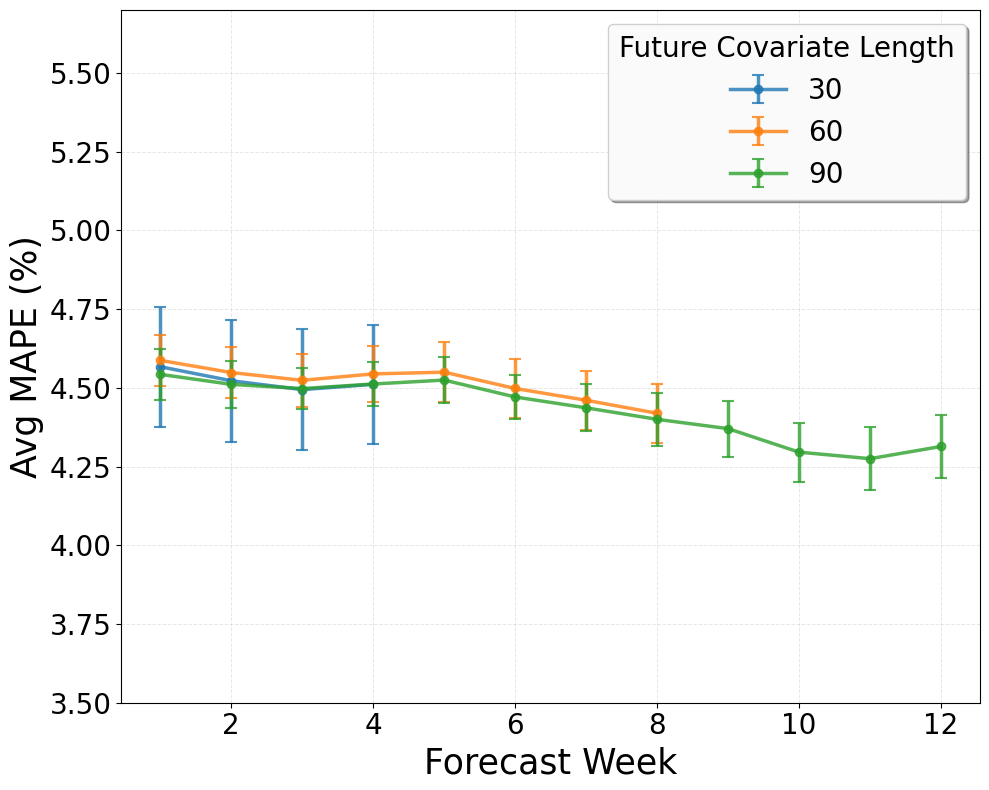} &
\includegraphics[width=0.22\textwidth]{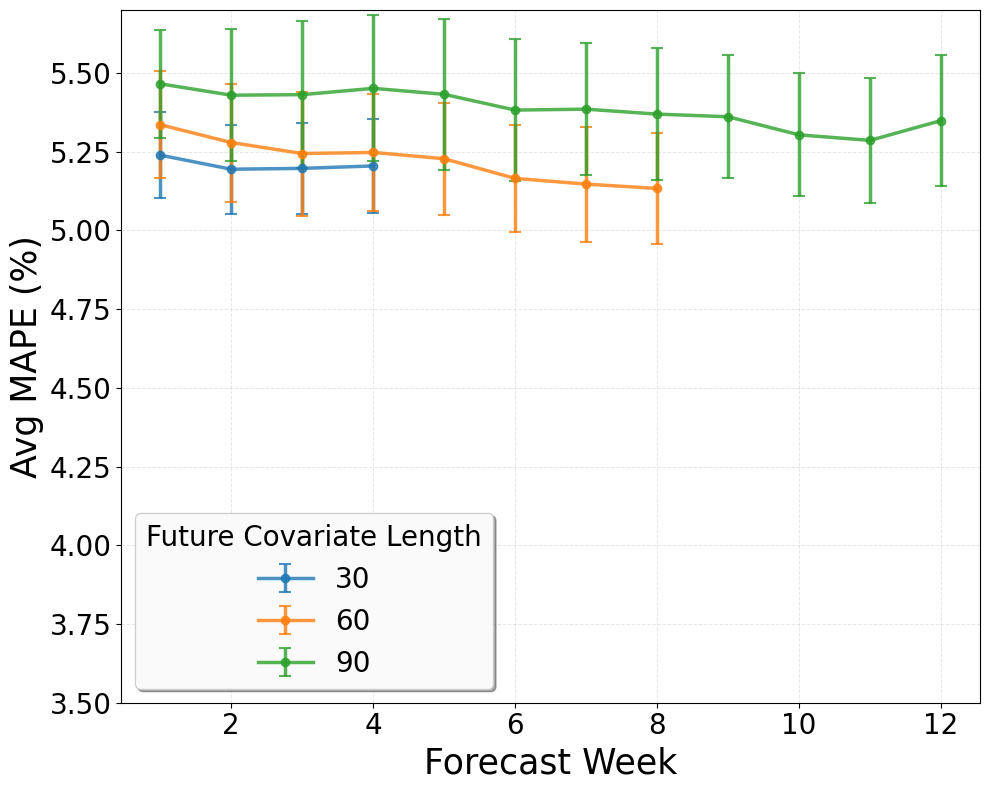} &
\includegraphics[width=0.22\textwidth]{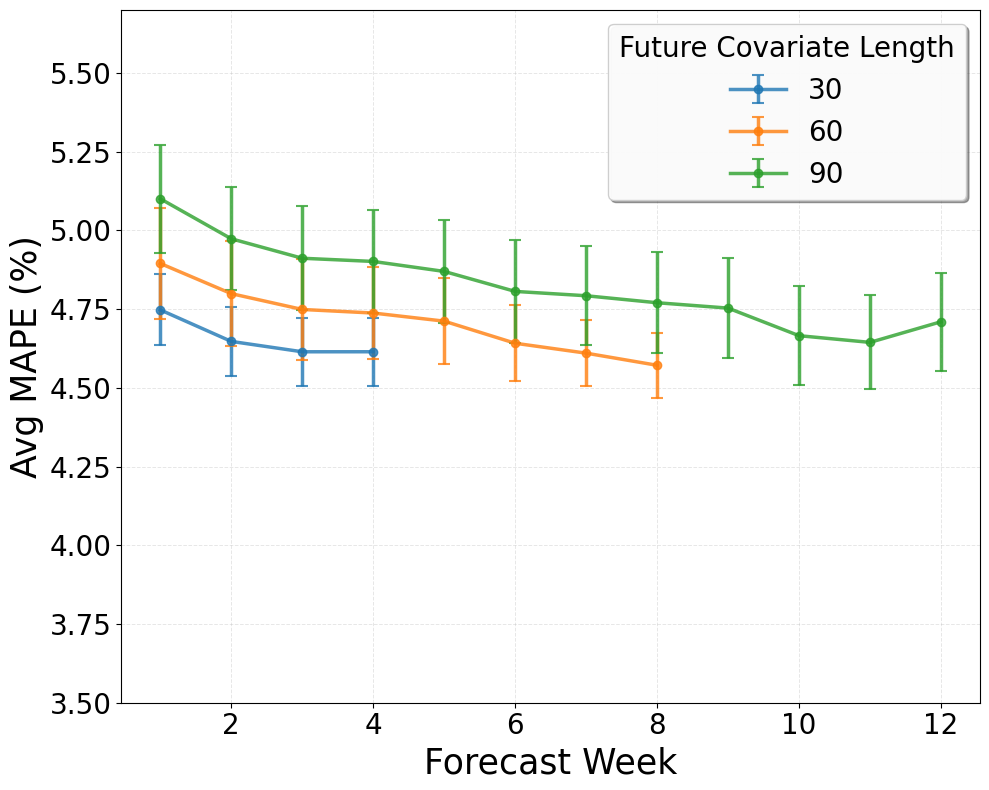} \\
\end{tabular}
\caption{Weekly forecast performance. Columns correspond to model types, and rows correspond to datasets. Each model is trained with forecasting horizons $H \in \{30,60,90\}$. The y-axis shows average MAPE by week (x-axis), and error bars denote the standard deviation across five random seeds.}
\label{fig:weekly_extended_cov}
\vspace{-3mm}
\end{figure*}
We systematically analyze how Chronos-2 (full fine-tuned) and MMMF models leverage recent historical information and future covariate availability during forecasting. To study this, we fine-tune models on task-specific datasets using different $H$. Figure~\ref{fig:weekly_extended_cov} reports the average weekly MAPEs of forecasts from models trained with different $H$ values. In the initial weeks, Chronos-2 significantly outperforms MMMF, indicating a stronger ability to exploit recent historical information for short-term forecasting. However, the weekly MAPE of Chronos-2 increases steadily with each subsequent week, eventually exceeds that of the MMMF models in later weeks. 

Furthermore, Chronos-2 models fine-tuned with larger $H$ consistently achieve lower weekly MAPEs than those trained with smaller $H$. This suggests that the model utilizes long-range future signals to refine its understanding of near-term trajectories, leading to performance gains even for the initial forecast periods. In contrast, MMMF-based LSTM and Transformer exhibits the opposite trend: models with access to longer future signals perform worse in the initial weeks than those with shorter future signals. This indicates that excessive masking degrades MMMF performance when the primary objective is accurate short-horizon forecasting. These results highlight that Chronos-2’s strength lies in leveraging recent history and long-range future signals for enhanced contextualization, leading to better short-term forecasting.

\subsection{Chronos-2's cold-start ICL capability}
\begin{figure*}[tbh]
    \centering
    \subfloat[]{
     \includegraphics[width=0.265\textwidth]{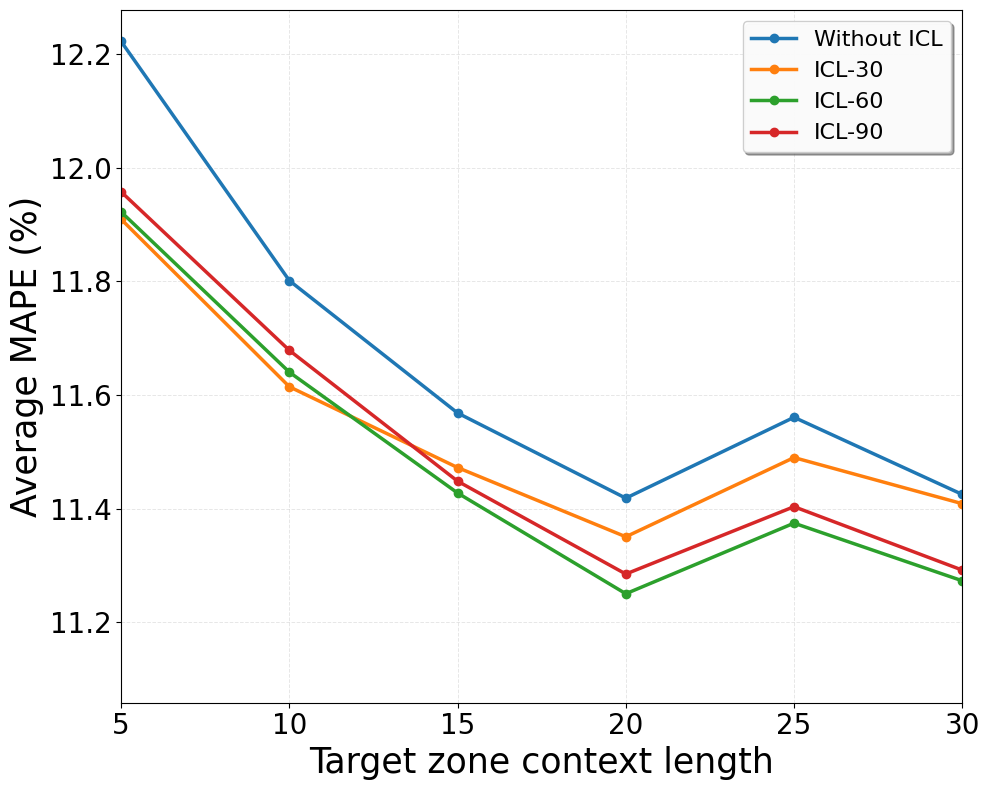} 
         \label{fig:icl_isone}
     }
    \subfloat[]{ 
           \includegraphics[width=0.265\textwidth]{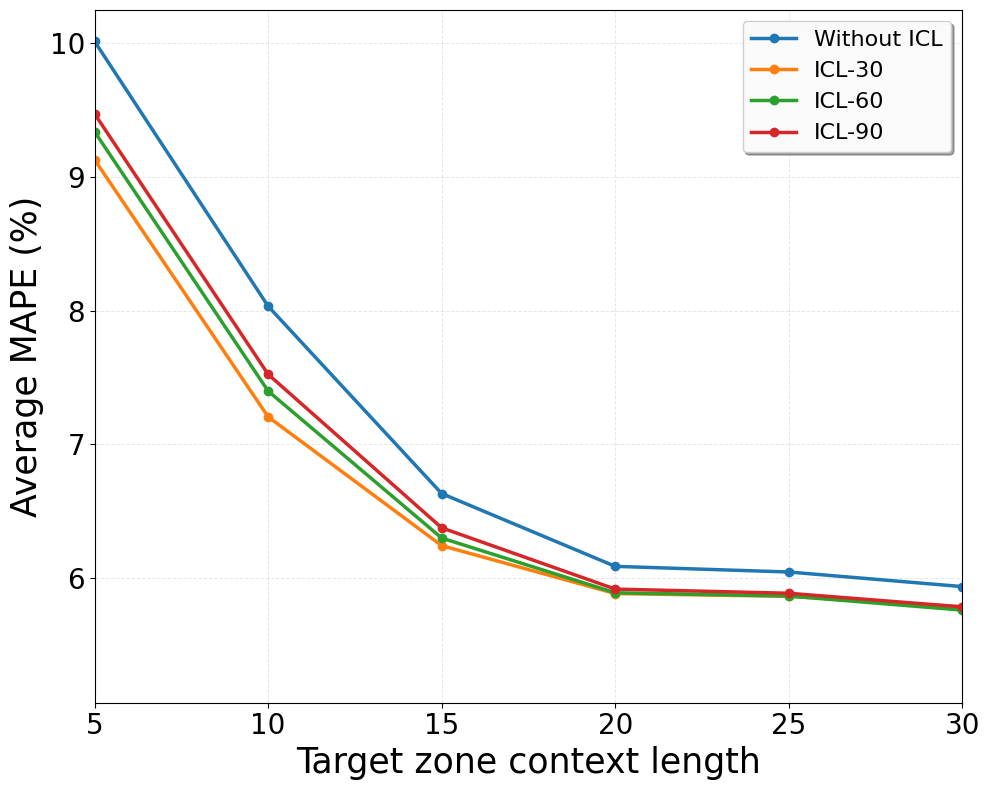}
          \label{fig:icl_entsoe}
          }  
    \caption{Forecast performance of zero-shot Chronos-2 in cold-start scenarios, with and without ICL in (a)~ISO New England (b)~ENTSO-E. X-axis represents the context length of each target zone. Each ICL line plot represents a cross-learning scenario with the context length of non-target zone $\in \{30, 60, 90\}$ Note: Y-axes are scaled differently across subplots to best visualize the trends within each dataset.}
    \label{fig:icl_cold_start}
    \vspace{-3mm}
\end{figure*}

We evaluate the in-context learning (ICL) capability of Chronos-2 in zero-shot, cold-start scenarios. For each target zone or country, we generate 30-day forecasts under two settings: (1) \emph{without ICL}, where only the target region’s load history of length $l_{\text{target}}$ is provided to the model, and (2) \emph{with ICL}, where, in addition to the target region’s history, the model is also given the historical load data from other regions, each of length $l_{\text{other}}$. To simulate cold-start scenario, we consider only cases where $l_{\text{other}} \geq l_{\text{target}}$. To analyze the impact of relative amount of historical information, we consider different values of $l_{\text{other}} \in \{30, 60, 90\}$ and vary $l_{\text{target}}$ from 5 to 30 days. The results are shown in Figure~\ref{fig:icl_cold_start}, where we report the average MAPE across all regions over the 30-day forecast horizon.

The results indicate that ICL leads to modest improvements in forecasting accuracy. The largest gains are observed in the most data-scarce setting ($l_{\text{target}} = 5$), with maximum MAPE reductions of approximately 0.3 and 1 percentage points for the ISO New England and ENTSO-E, respectively. However, these improvements diminish as the target-region history increases. When $l_{\text{target}} = 30$, the best achieved MAPEs of 11.27 (ISO New England, $l_{\text{other}} = 60$) and 5.8 (ENTSO-E, $l_{\text{other}} = 60$) remain higher than those obtained by the best MMMF models (10.64 and 5.66, respectively) and by the fine-tuned Chronos-2 (10.07 and 5.16, respectively) for the same 30-day forecasting task without covariates. These results suggest that while Chronos-2 exhibits some degree of cross-series transfer in cold-start settings, the practical impact of its ICL capability is limited in these forecasting tasks.

\subsection{Computational and operational implications}
Adopting foundation models like Chronos-2 in operational power system settings introduces significant trade-offs regarding computational cost and model lifecycle management.

\textbf{Inference Time:} 
There is a stark contrast in model complexity. Chronos‑2 (120M parameters) has approximately three orders of magnitude more parameters than the task‑specific baselines (e.g., MMMF‑Transformer with 100K parameters). Consequently, Chronos‑2's inference time (3.1 ms per sample) is nearly two orders of magnitude longer than MMMF‑Transformer's (0.044 ms per sample) even on an NVIDIA RTX 1000 Ada Generation Laptop GPU. For real‑time applications requiring frequent updates across thousands of nodes (e.g., feeder‑level forecasting), the computational overhead of foundation models may be prohibitive without substantial hardware acceleration.

\textbf{Model Hosting and Management:} Fine-tuning foundation models for utility-scale applications presents unique hosting challenges. Utilities often require distinct forecast models for hundreds or thousands of individual zones, substations, or feeders. Storing and serving a 120M-parameter model for each entity is impractical due to storage costs and GPU memory constraints. This highlights the operational necessity of parameter-efficient fine-tuning methods like LoRA. By storing a single frozen backbone model and swapping only the lightweight adapter weights (typically $<1\%$ of the total parameters) for each target zone, utilities can mitigate the storage explosion.

In summary, while foundation models offer superior accuracy for short-horizon planning tasks where latency is less critical, lightweight task-specific models remain operationally advantageous for high-frequency, high-volume real-time dispatch environments.

\section{Conclusions}\label{sec:conclude}
Load forecasting is an important problem in power system operations, and the recent interest in time-series foundation models has been driven by the promise to reduce the reliance on task-specific model development while maintaining strong performance on downstream tasks, such as forecasting. This study focused on rigorously evaluating the performance of Chronos-2 on real-world grid load forecasting, by comparing it against established task-specific deep learning models.

To this end, we considered two real-world utility datasets (ISO New England and ENTSO-E), consisting of electricity load data from multiple geographical regions, calendar features, and diverse weather covariates. Our results show that zero-shot Chronos-2 is unable to match the performance of the best task-specific models. Task-specific fine-tuning enables Chronos-2 to achieve superior performance at short forecast horizons. However, its forecasting errors increase more rapidly with horizon length and ultimately exceed those of task-specific models at longer horizons. We find that full fine-tuning consistently outperforms parameter-efficient adaptation via LoRA, particularly when covariates are available. In addition, explicitly incorporating calendar features through cyclical encoding during fine-tuning leads to noticeable improvements in forecasting accuracy. Finally, our analysis of cold-start scenarios indicates that Chronos-2’s in-context learning provides only marginal benefits, limiting its practical impact in the load forecasting tasks considered in this study.

Overall, this study contributes practical insights into the applicability, strengths, and limitations of emerging foundation models in power grid load forecasting. Future work will extend this evaluation to other foundation models, higher-resolution forecasting tasks such as hourly week-ahead forecasting and feeder-level load prediction, to assess scalability. We also plan to investigate the probabilistic forecasting capabilities of these models and compare quantile predictions against point predictions for risk-sensitive operational settings.

\bibliographystyle{IEEEtran}
\bibliography{References}

\end{document}